### ARTIFICIAL INTELLIGENCE

# AADS: Augmented autonomous driving simulation using data-driven algorithms

W. Li[1,2,3]*[†], C. W. Pan[4,5]*, R. Zhang[6]*, J. P. Ren[6], Y. X. Ma[7], J. Fang[1,2], F. L. Yan[1,2], Q. C. Geng[8], X. Y. Huang[1,2], H. J. Gong[3], W. W. Xu[6], G. P. Wang[4], D. Manocha[9][†], R. G. Yang[1,2][†]

Simulation systems have become essential to the development and validation of autonomous driving (AD) technologies. The prevailing state-of-the-art approach for simulation uses game engines or high-fidelity computer graphics (CG) models to create driving scenarios. However, creating CG models and vehicle movements (the assets for simulation) remain manual tasks that can be costly and time consuming. In addition, CG images still lack the richness and authenticity of real-world images, and using CG images for training leads to degraded performance. Here, we present our augmented autonomous driving simulation (AADS). Our formulation augmented real-world pictures with a simulated traffic flow to create photorealistic simulation images and renderings. More specifically, we used LiDAR and cameras to scan street scenes. From the acquired trajectory data, we generated plausible traffic flows for cars and pedestrians and composed them into the background. The composite images could be resynthesized with different viewpoints and sensor models (camera or LiDAR). The resulting images are photorealistic, fully annotated, and ready for training and testing of AD systems from perception to planning. We explain our system design and validate our algorithms with a number of AD tasks from detection to segmentation and predictions. Compared with traditional approaches, our method offers scalability and realism. Scalability is particularly important for AD simulations, and we believe that real-world complexity and diversity cannot be realistically captured in a virtual environment. Our augmented approach combines the flexibility of a virtual environment (e.g., vehicle movements) with the richness of the real world to allow effective simulation.

## INTRODUCTION

Autonomous vehicles (AVs) have attracted considerable attention in recent years from researchers, venture capitalists, and the general public. The societal benefits in terms of safety, mobility, and environmental concerns are expected to be tremendous and have captivated the attention of people across the globe. However, in light of recent accidents involving AVs, it has become clear that there is still a long way to go to meet the high standards and expectations associated with AVs.

Safety is the key requirement for AVs. It has been argued that an AV has to be test-driven hundreds of millions of miles in challenging conditions to demonstrate statistical reliability in terms of reductions in fatalities and injuries (1), which could take tens of years of road tests even under the most aggressive evaluation schemes. New methods and metrics are being developed to validate the safety of AVs. One possible solution is to use simulation systems, which are common in other domains such as law enforcement, defense, and medical training. Simulations of autonomous driving (AD) can serve two purposes. The first is to test and validate the capability of AVs in terms of environmental perception, navigation, and control. The second is to generate a large amount of labeled training data to train machine learning methods, e.g., a deep neural network. The second purpose has recently been adopted in computer vision (2, 3).

The most common way to generate such a simulator is to use a combination of computer graphics (CG), physics-based modeling, and robot motion planning techniques to create a synthetic environment in which moving vehicles can be animated and rendered. A number of simulators have recently been developed, such as Intel's CARLA (4), Microsoft's AirSim (5), NVIDIA's Drive Constellation (6), Google/Waymo's CarCraft (7), etc.

Although all of these simulators achieve state-of-the-art synthetic rendering results, these approaches are difficult to deploy in the real world. A major hurdle is the need for high-fidelity environmental models. The cost of creating life-like CG models is prohibitively high. Consequently, synthetic images from these simulators have a distinct, CG-rendered look and feel, i.e., gaming or virtual reality (VR) system quality. In addition, the animation of moving obstacles, such as cars and pedestrians, is usually scripted and lacks the flexibility and realism of real scenes. Moreover, these systems are unable to generate different scenarios composed of vehicles, pedestrians, or bicycles, as observed in urban environments.

Here, we present a data-driven approach for end-to-end simulation for AD: augmented autonomous driving simulation (AADS). Our method augments real-world pictures with a simulated traffic flow to create photorealistic simulation scenarios that resemble real-world renderings. Figure 1 shows the pipeline of our AADS system and its major inputs and outputs. Specifically, we proposed using light detection and ranging (LiDAR) and cameras to scan street scenes. We decomposed the input data into background, scene illumination, and foreground objects. We presented a view synthesis technique to enable changing viewpoints on the static background. The foreground vehicles were fitted with three-dimensional (3D) CG models. With accurately estimated outdoor illumination, the 3D vehicle models, computer-generated pedestrians, and other movable subjects could be repositioned and rendered back into the background images to create photorealistic street view images that looked like they were captured from a dashboard camera on a vehicle. Furthermore, the simulated traffic flows, e.g., the placement and movement of synthetic objects, were based on captured real-world vehicle trajectories that looked natural and captured the complexity and diversity of real-world scenarios.

[1]Baidu Research, Beijing, China. [2]National Engineering Laboratory of Deep Learning Technology and Application, Beijing, China. [3]Nanjing University of Aeronautics and Astronautics, Nanjing, China. [4]Beijing Engineering Technology Research Center of Virtual Simulation and Visualization, Peking University, Beijing, China. [5]Deepwise AI Lab, Beijing, China. [6]Zhejiang University, Hangzhou, China. [7]University of Hong Kong, Hong Kong, China. [8]Beihang University, Beijing, China. [9]University of Maryland, College Park, MD, USA.
*These authors contributed equally to this work.
†Corresponding author. Email: liwei87@baidu.com (W.L.); yangruigang@baidu.com (R.G.Y.); dm@cs.umd.edu (D.M.)

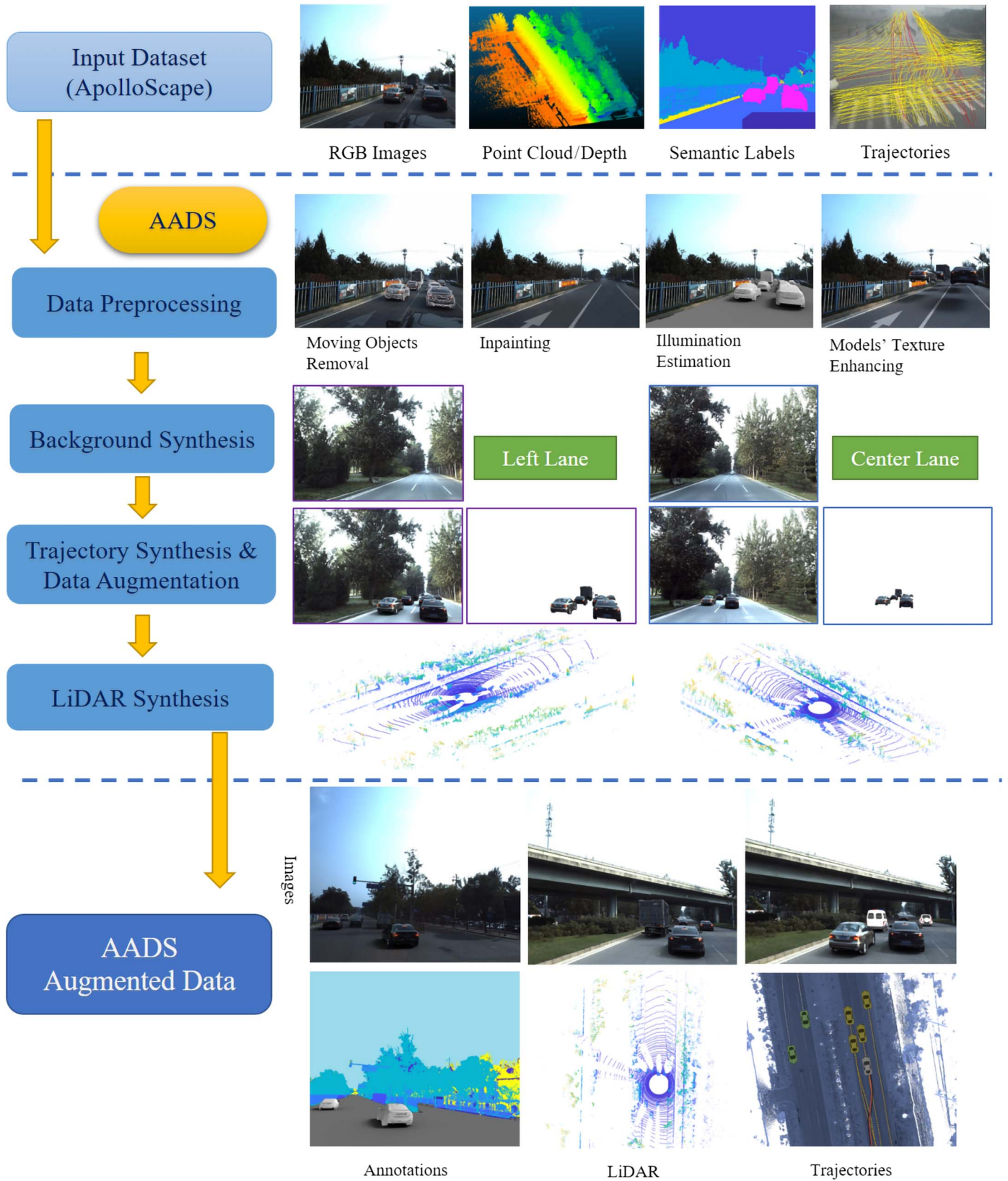

**Fig. 1. The inputs, processing pipeline, and outputs of our AADS system. Top:** The input dataset. **Middle:** The pipeline of AADS is shown between the dashed lines and contains data preprocessing, novel background synthesis, trajectory synthesis, moving objects' augmentation, and LiDAR simulation. **Bottom:** The outputs from the AADS system, which include synthesized RGB images, a LiDAR point cloud, and trajectories with ground truth annotations.

Compared with traditional VR-based or game engine–based AV simulation systems, AADS provides more accurate end-to-end simulation capability without requiring costly CG models or tedious programming to define the traffic flow. Therefore, it can be deployed for large-scale use, including training and evaluation of new navigation strategies for the ego vehicle.

The key to AADS's success is the wide availability of 3D scene scans and vehicle trajectory data, both of which are needed for automatic generation of new traffic scenarios. We will also release part of the real-world data that we have collected for the development and evaluation of AADS. The data were fully annotated by a professional labeling service. In addition to AADS, they may also be used for many perception- and planning-related tasks to drive further research in this area.

This paper includes the following technological advances:

1) A data-driven approach for AD simulation: By using scanned street view images and real trajectories, both photorealistic images and plausible movement patterns can be synthesized automatically. This direct scan-to-simulation pipeline, with little manual intervention, enables large-scale testing of autonomous cars virtually anywhere and anytime within a closed-loop simulation.

2) A view synthesis method to enable view interpolation and extrapolation with only a few images: Compared to previous approaches, it generates better quality images with fewer artifacts.

3) A new set of datasets, including the largest set of traffic trajectories and the largest 3D street view dataset with pixel/point level annotation: All of these are captured in metropolitan areas, with dense and complex traffic patterns. This kind of dense urban traffic poses notable challenges for AD.

## Previous methods

Simulation for AD is a very large topic. Traditionally, simulation capabilities have been primarily used in the planning and control phase of AD, e.g., (8–14). More recently, simulation has been used in the entire AD pipeline, from perception and planning to control [see the survey by Pendleton et al. (15)].

Although Waymo has claimed that its AV has been tested for billions of miles in their proprietary simulation system, CarCraft (7), little technical detail has been released to the public in terms of its fidelity for training machine learning methods. Researchers have tried to use images from video games to train deep learning–based perception systems (16, 17).

Recently, a number of high-fidelity simulators dedicated to AD simulation have been developed, such as Intel's CARLA (4), Microsoft's AirSim (5), and NVIDIA's Drive Constellation (6). They allow end-to-end, closed-loop training and testing of the entire AD pipeline beyond the generation of annotated training data. All of these simulators have their basis in current gaming techniques or engines, which generate high-quality synthetic images in real time. A limitation of these systems is the fidelity of the resulting environmental model. Even with state-of-the-art rendering capabilities, the images produced by these simulators are obviously synthetic. Current state-of-the-art CG rendering may not provide enough accuracy and details for machine learning methods.

With the availability of LiDAR devices and advances in structure from motion, it is now possible to capture large urban scenes in 3D. However, turning the large-scale point cloud (PC) into a CG quality–rendered image is still an ongoing research problem. Models reconstructed from these point clouds often lack details or complete textures (18). In addition, AD simulators have to address the problem of realistic traffic patterns and movements. Traditional traffic flow simulation algorithms mainly focus on generating trajectories for vehicles and do not take into account the realistic movements of individual cars or pedestrians. One of the challenges is to simulate realistic traffic patterns, particularly in complex situations, when traffic is dense and involves heterogenous agents (e.g., an intersection scenario with pedestrians in a crosswalk).

Our work is related to the approach described by Alhaija et al. (19) in which 3D vehicle models were rendered onto existing captured real-world background images. However, the observation viewpoint was fixed at capture time, and the 3D models were chosen from an existing 3D repository that may or may not match those in the real-world images. Their approach can be used to augment still images for training perception applications. In contrast, with the ability to freely change the observation viewpoint, our system could not only play a role in data augmentation but also enhance a closed-loop simulator such as CARLA (4) or AirSim (5). Further enhanced by realistic traffic simulation ability, our system can also be used for path planning and driving decision applications. In those dynamic applications, our system can generate data in a loop for reinforcement learning and learning-by-demonstration algorithms. Overall, the proposed approach enables closed-loop, end-to-end simulation without the need for environmental modeling and human intervention.

## RESULTS

Because AADS is data driven, we first explain the datasets that have been collected. Some of the datasets have already been released, and others will be released with this paper. We then show results for the synthesis of virtual views and generation of traffic flows, two key components of AADS. Last, we evaluated AADS's effectiveness for AD simulation. Specifically, we show that the simulated red-green-blue (RGB) and LiDAR images were useful for improving the performance of the perception system, whereas the simulated trajectories were useful for improving predictions of obstacle movements—a critical component for the planning and control phases for autonomous cars.

### The datasets

When collecting a dataset, we used a hardware system consisting of two RIEGL laser scanners, one real-time line-scanning LiDAR (Velodyne 64-line), a VMX-CS6 stereo camera system, and a high-precision inertial measurement unit (IMU)/global navigation satellite system (GNNS). With the RIEGL scanners, our system could obtain higher-density point clouds with better accuracy than widely used LiDAR scanners, whereas the VMX-CS6 system provided a wide baseline stereo camera with high resolution (3384 by 2710). With the Velodyne LiDAR, we could obtain the shapes and positions of moving objects. To scan a scene, the hardware was calibrated, synchronized, and then mounted on the top of a mid-size sport utility vehicle (SUV) that cruised around the target scene at an average speed of 30 km/hour. Note that the RGB images were taken about once every meter.

Instead of fully annotating all 2D/RGB and 3D/point cloud data manually, we developed a labeling pipeline to make our labeling process accurate and efficient. Because 2D labeling is expensive in terms of time and labor, we combined the two stages, i.e., 3D labeling and 2D labeling.

By using easy-to-label 3D annotations, we could automatically generate high-quality 2D annotations of static backgrounds/objects in all the image frames by 3D-2D projections. Details of the labeling process can be found in (20).

For each image frame, we annotated 25 different classes covered by five groups in both 3D point clouds. In addition to standard annotation classes, such as cars, motorcycles, traffic cones, and so on, we added a new "tricycle" class, a popular mode of transportation in East Asian countries. We also annotated 35 different lane markings in both 2D and 3D that were not previously available in open datasets. These lane markings were defined on the basis of color (e.g., white and yellow), type (e.g., solid and broken), and usage (e.g., dividing, guiding, stopping, and parking).

The table in Fig. 2 compares our dataset and other street view datasets. Our dataset outperformed other datasets in many aspects, such as scene complexity, number of pixel-level annotations, number of classes, and so on. We have released 143,906 video frames and corresponding pixel-level annotations. Images were assigned to three degrees of difficulty (e.g., easy, moderate, and hard) based on scene complexity, a measure of the number of movable objects in an image. Our dataset also contains challenging lighting conditions, such as high-contrast regions due to sunlight and shadows from overpasses. We named the dataset of RGB images ApolloScape-RGB. We also provided 3D point-level annotations in ApolloScape point cloud (ApolloScape-PC) dataset, which are not available in other street view datasets.

In addition to this article, we also announce ApolloScape-TRAJ, a large-scale dataset for urban streets that includes RGB image sequences and trajectory files. It is focused on trajectories of heterogeneous traffic agents for planning, prediction, and simulation tasks. The dataset includes RGB videos with around 100,000 images with a resolution of 1920 by 1080 and 1000 km of trajectories for all kinds of moving traffic agents. We used the Apollo acquisition car to collect traffic data and generate trajectories. In Beijing, we collected a dataset of trajectories under a variety of lighting conditions and traffic densities. The dataset includes many challenging scenarios involving many vehicles, bicycles, and pedestrians moving around one another.

### Evaluations of augmented background synthesis

An important part of our AADS system is synthesizing background images in specific views using images captured in fixed views when running closed-loop simulations. This ability stems from the utilization of the image-based rendering technique and avoids prerequisite modeling of the full environment.

There is a large literature on image-based rendering techniques, although relatively little has been written on capturing scenes with sparse images. We focused on wide baseline stereo image–based rendering for street view scenes: The overlap between left images and right images may be less than half the size of full images. Technically, obtaining reliable depth is an important challenge for image-based rendering techniques. Thus, methods such as (21) use the multiview stereo method to estimate depth maps. However, most street view datasets provide laser-scanned point clouds, which can be used to generate initial depth maps by rendering point clouds. As point clouds tend to be sparse and noisy, initial estimates of depth maps are full of outliers and holes and need to be refined before they are passed on to downstream processing. Thus, we proposed an effective depth refinement method that included depth filtering and completion procedures. To evaluate our depth refinement method, we used initial and refined depth maps (Fig. 3, B and E) to synthesize the same novel view. Results are shown in Fig. 3 (F and G, respectively).

When using depth maps without refinement to run image-based rendering, the results suffered from artifacts near errors and holes in depth maps. Specifically, in Fig. 3 (F and H), fluctuations appeared in the green rectangle as the view changed, whereas window frames were kept straight when using refined depth in the yellow rectangle.

To evaluate our image-based rendering algorithm (specifically the novel view synthesis algorithm) with refined depth maps, we compared our method with two representative approaches: the content preserving warping method by Liu *et al.* (22) and the method by Chaurasia *et al.* (23). Note that, in the implementation of the method by Chaurasia *et al.* (23), we used the similarity of super pixels (24) to complete the depth map and perform a local shape-preserving warp on each super pixel.

The synthesized images in Fig. 3 were generated using four reference images. Because images were captured by a stereo camera, the four reference images could be considered as two pairs of stereo images with close to parallel views in which the angle between two optical axes of the stereo images is small, but the baseline is relatively wide (about 1 m). We compared our view interpolation and extrapolation results with classical methods. As shown in the third row of Fig. 3, the method by Liu *et al.* (22) performed well for small changes in the novel view compared to the input views. When the view translation became larger, view distortion artifacts became apparent (such as the fence in the green rectangle, the shape of which is deformed inappropriately). For the method by Chaurasia *et al.* (23), ghost artifacts appeared when neighboring super pixels were assigned to inappropriate or incorrect depths. Our method obtained correct depths and preserved invariant shapes of objects when the view changes, handling both interpolation and extrapolation. The fourth row of Fig. 3 evaluates another scene with both a wide baseline and a large rotation angle. Because of large changes in the novel view, neither the method by Liu *et al.* (22) nor the method by Chaurasia *et al.* (23) aligned well with neighboring reference views. As shown in the figure, curbstones in the green rectangle and the white lane marker in the yellow rectangle reveal misalignment artifacts. In addition, because of tone inconsistencies in the input images, seams are prominent in the results of the methods by Liu *et al.* (22) and Chaurasia *et al.* (23). In contrast, our method could effectively eliminate misalignment and seam artifacts.

To further illustrate the effectiveness of our view synthesis approach for closed-loop simulation, we have included a video (movie S3) that shows the synthesized front camera view from a driving car that changes lanes several times. Our view synthesis approach is sufficient for handling such lane changes because it interpolates or extrapolates the viewpoint.

### Evaluations of trajectories synthesis

Another pillar for AADS is its ability to generate plausible traffic flow, particularly when there are interactions between vehicles and pedestrians, e.g., heterogeneous agents who move at different speeds and with different dynamics. This topic is a full research area in its own right, and we developed techniques for heterogeneous agent simulations. For the sake of completeness, we briefly show the main result in Fig. 4. Readers are referred to (25) for more technical details. Specifically, Fig. 4 shows the comparison with the ground truth from the input dataset, results of our simulation method, and results of the method by Chao *et al.* (26), a state-of-the-art multiagent simulation approach. In the evaluation, the traffic was simulated on a straight four-lane road. For our method, the number, positions, and velocities of agents were randomly initialized according to the dataset. We evaluated the comparison using the metric of velocity and minimum distance probability distributions. The metrics are divided into 30 intervals. The probability of each interval is the divisor of the sample number in this interval and the total sample number. As shown in Fig. 4, our simulation results are closer to the input data in both the velocity distribution and the minimum distance distribution.

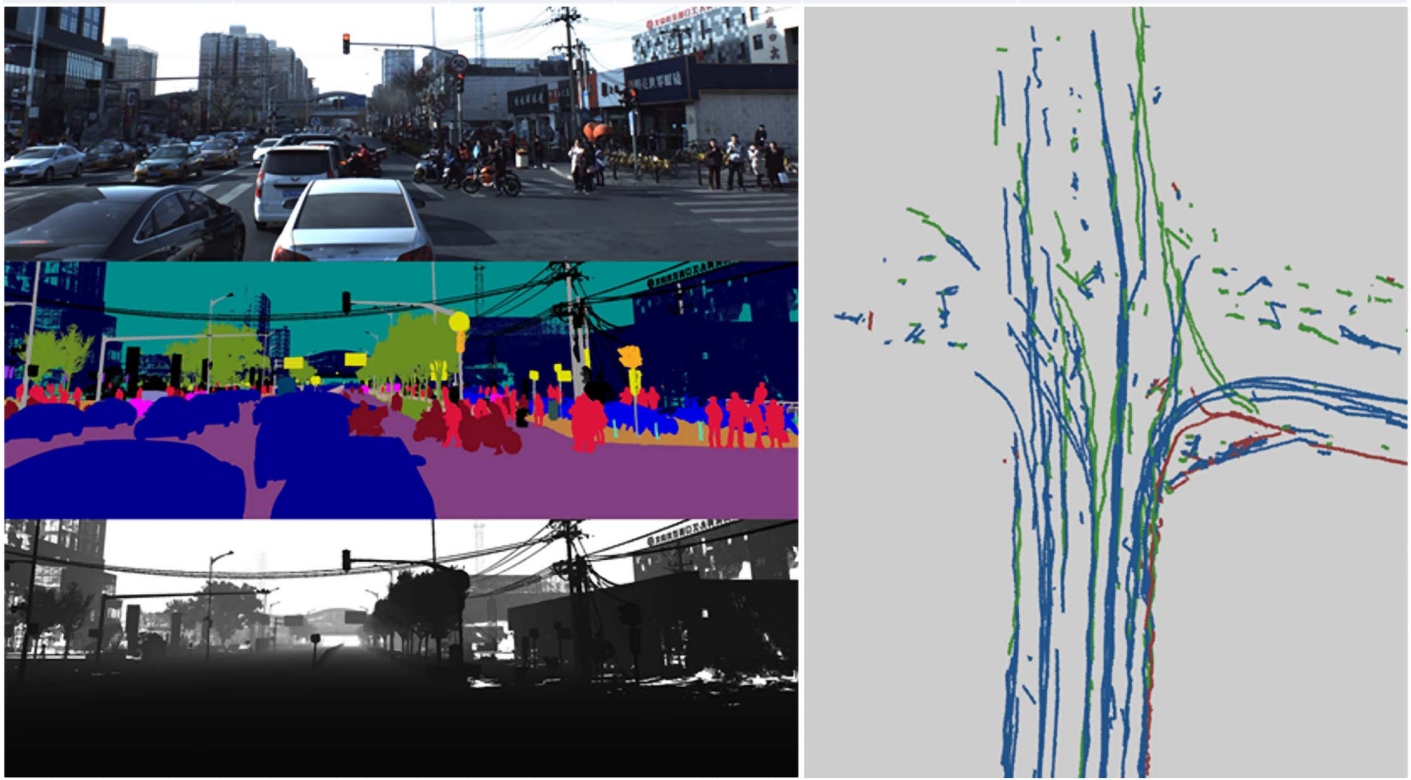

|  | KITTI | CityScapes | Mapillary | BDD100K | ApolloScape | | | |
|---|---|---|---|---|---|---|---|---|
| Total images | 14,999 | 25,000 | 25,000 | 120,000,000 | 143,906 | | | |
| Annotated images (bounding-box level) | 14,999 | no | no | 100,000 | no | | | |
| Annotated images (pixel level) | 400 | 5,000 fine 20,000 coarse | 25,000 | 10,000 | 143,906 | | | |
| Scene Complexity (average per image) | bounding-box level | pixel level | pixel level | bounding-box level | pixel level | | | |
|  | person: 0.8 vehicle: 4.1 | person: 7.0 vehicle: 11.8 | - | person: ~1.3 vehicle: ~11.0 | difficulty | easy | moderate | hard |
|  |  |  |  |  | person | 1.1 | 6.2 | 16.9 |
|  |  |  |  |  | vehicle | 12.7 | 24.0 | 38.1 |
| Diversity | day time | day time 50 cities | various weather day & night 6 continents | various weather day & night 4 regions in US | various weather day time 4 regions in China | | | |
| 3D Annotation | box-level | no | no | no | point-level | | | |
| Video Annotation | box-level | no | no | no | Pixel-level | | | |
| Lane Annotation | no | no | 2D/2 classes | 2D/8 classes | 3D/2D video 28 classes | | | |
| Location Accuracy | cm | - | meter | meter | cm | | | |

**Fig. 2. The ApolloScape dataset and its extension. Top:** Table comparing ApolloScape with other popular datasets. **Bottom:** RGB images, annotations, and a point cloud from top to bottom (left) and some labeled traffic trajectories from the dataset (right).

### AADS evaluations by AD applications
As shown in Fig. 1, simulation and our AADS can simultaneously produce the following augmented data: (i) photorealistic RGB images with annotation information such as semantic labels, 3D bounding boxes, etc.; (ii) an augmented LiDAR point cloud; and (iii) typical traffic flows. In the following evaluations, those data augmentations were synthesized on the basis of our ApolloScape dataset. We summarize the AADS synthetic data and evaluations in terms of RGB images, point clouds, and trajectories as follows:

1) AADS-RGB: For the baseline training set of ApolloScape-RGB, we augmented RGB images with AADS and generated corresponding annotations for augmented moving agents. This dataset is named AADS-RGB and was used to evaluate our image synthesis method.

2) AADS-PC: With our AADS system, we synthesized up to 100,000 new point cloud frames by simulating the Velodyne HDL-64E S3 LiDAR sensor based on the ApolloScape-PC dataset. The simulation dataset has the same object categories as and numbers of objects in each category similar to ApolloScape-PC.

3) AADS-TRAJ: Our AADS system can also produce new trajectories based on the ApolloScape-TRAJ dataset. We further evaluated such augmented data using a trajectory prediction method.

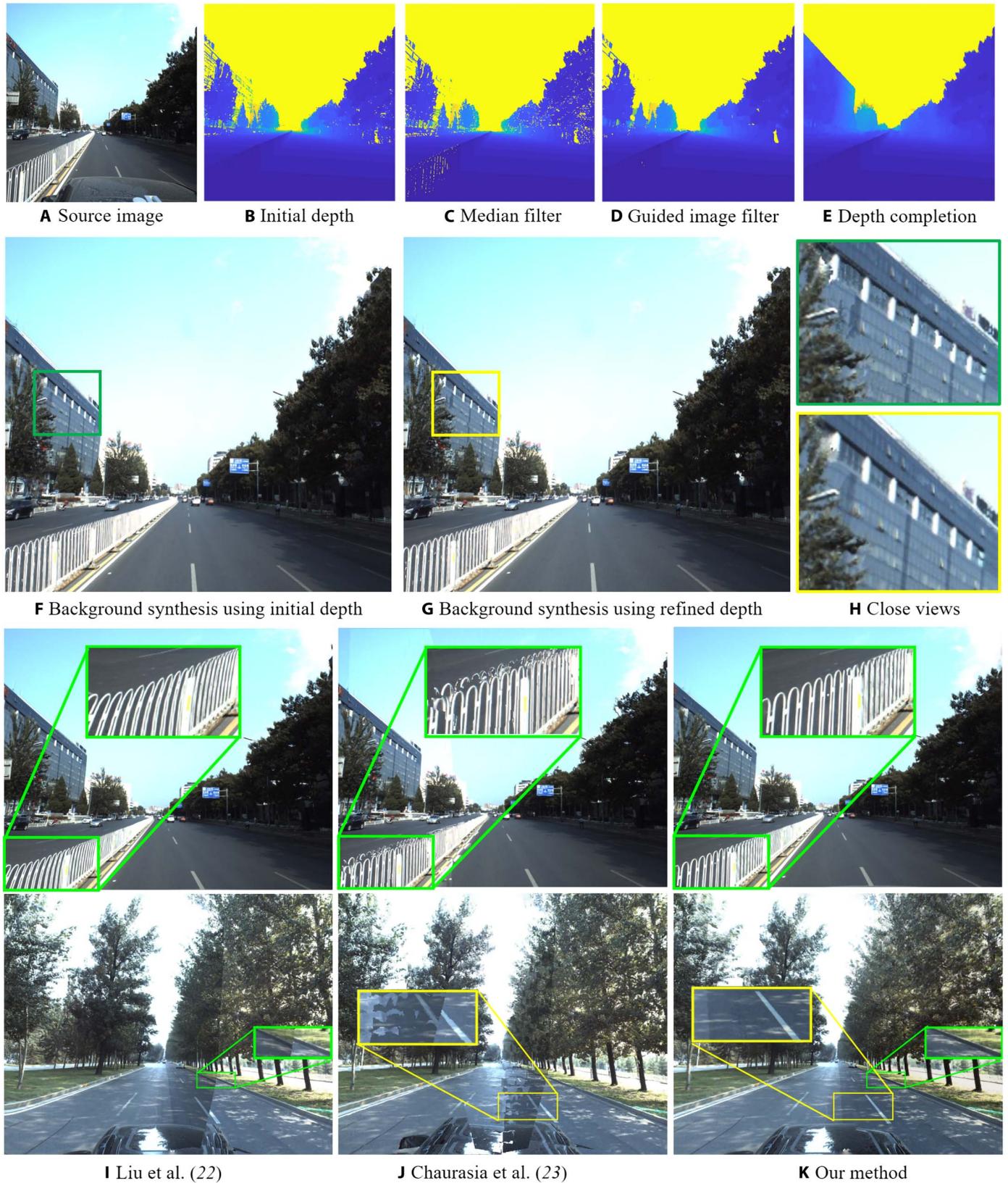

**Fig. 3. View synthesis results and effectiveness of depth refinement.** (**A** and **B**) Raw RGB and depth images in our dataset, respectively. (**C** to **E**) Results of depth refinement after filtering and completion. (**F** and **G**) Results of view synthesis using initial and refined depths with close views in (**H**). (**I** to **K**) Final results of view synthesis using the method by Liu *et al.* (*22*), the method by Chaurasia *et al.* (*23*), and our method, respectively.

**Object detection with AADS-RGB**

For the evaluations of AADS' capability to simulate camera images, we used two real and three virtual datasets: ApolloScape-RGB–annotated images (ApolloScape-RGB), CityScapes, virtual KITTI (VKITTI), synthesized data from the popular simulator CARLA, and our synthesized data (AADS-RGB).

We used the VKITTI (2) dataset to compare our system with a fully synthetic method. The full dataset contains 21,260 images with different weather and lighting conditions. A total of 1600 images were randomly selected as a training set.

CityScapes (27) is a dataset of urban street scenes. There are 5000 annotated images with fine instance-level semantic labels. We used the validation set of 492 images as the testing dataset.

CARLA (4) is the most recent and popular VR simulator for AD. Up to now, it provides two manually built scenes with car models. Because the size of the scene is limited, we generated 1600 images distributed as evenly as possible in the simulated scene.

In this section, we show the effectiveness of the AADS-RGB data. We used the state-of-the-art objection detection algorithm Mask R-CNN (28) to perform the experiments. The results were compared with the standard average precision metric of an intersection over union (IoU) threshold of 50% (AP50) and 70% (AP70) and a mean bounding box AP (mAP) with an across threshold at IoU ranging from 5 to 95% in steps of 5%. Because we mainly augmented textured vehicles onto images in our object detection evaluation, the evaluation results came from vehicles.

Synthetic data generation is an easy way to obtain large-scale datasets and has been proven to be effective in AD. However, the data statistics and distribution limit the capabilities of virtual data. When applying a model trained with synthetic data to real images, there is a domain gap. Because our simulation method was built on realistic background, placement, and moving object synthesis, it effectively reduced the domain problem. Our method produced an image (Fig. 5C) that is more visually similar to a real image from CityScapes (Fig. 5D) than it is to the VR simulator CARLA (Fig. 5A) or the fully synthetic dataset VKITTI (Fig. 5B), i.e., images from our system have small domain gaps.

To quantitatively verify the effectiveness of our simulated data, we chose to train object detectors with our data and data from CARLA and VKITTI. The trained detectors were tested on the CityScape dataset, which has no overlap with any of the training sets.

We trained models on CARLA-1600, VKITTI-1600, ApolloScape-RGB-1600, and AADS-RGB-2400 separately, where the suffix shows the number of images used for training. Then, the object detection performance of the trained model was evaluated on the CityScapes validation set. Results are shown in Fig. 5 (right). It can be seen that, because of the domain gap, the metrics of ApolloScape-1600 are higher than those of VKITTI-1600 or CARLA-1600. Note that images in VKITTI are smaller than images in other datasets. We therefore applied the VKITTI-1600 model on downsampled CityScapes to make the comparisons fair. Otherwise, the VKITTI-1600 model tended to miss large cars, leading to a degradation in detection performance. Adding

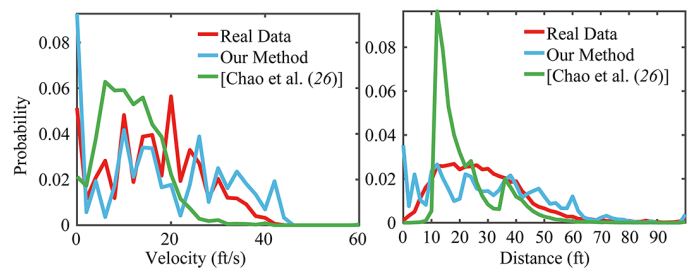

**Fig. 4. Comparison of traffic synthesis.** Velocity and minimum distance distribution of traffic simulation using our method, the method by Chao *et al.* (26), and the ground truth.

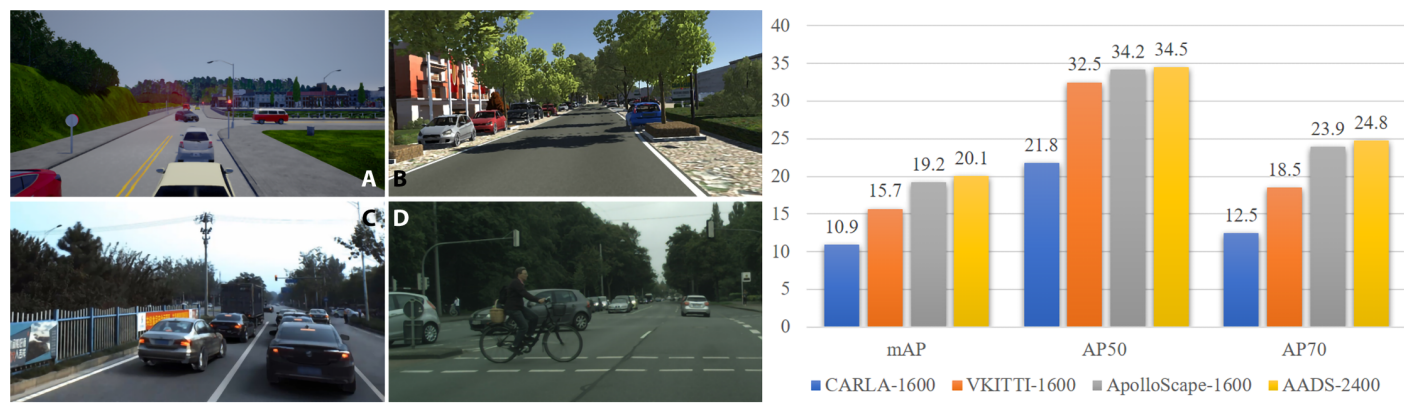

**Fig. 5. RGB image augmentation evaluations.** The four images on the left were selected from CARLA (**A**), VKITTI dataset (**B**), our AADS-RGB dataset (**C**), and the testing dataset CityScapes (**D**). The bar graph on the right shows the evaluation results using mAP, AP50, and AP70 metrics, respectively.

800 simulated images to ApolloScape-1600 (AADS-RGB-2400) improved the results by roughly 1%. This demonstrates that our simulation data may be closer to real-world data than data from VR.

**Instance segmentation with AADS-PC**

To evaluate AADS-PC simulations, we used the KITTI-PC dataset, the ApolloScape-PC, and our simulated point cloud (AADS-PC).

The KITTI-PC dataset (29) consists of 7481 training and 7518 testing frames. These real point cloud frames were labeled corresponding to captured RGB images in the front view. This dataset provides evaluation benchmarks for (i) 2D object detection and orientation estimation, (ii) 3D object detection, and (iii) bird's eye view evaluations.

On the basis of those datasets, we evaluated our AADS system using 3D instance segmentation. It was a typical point cloud–based AD application, which simultaneously ran 3D object detection and point cloud segmentation. We used the state-of-the-art algorithm PointNet++ (30) to perform quantitative evaluation. The results were evaluated using a mAP named mAP(Bbox) and a mean mask AP named mAP(mask).

We evaluated the accuracy and effectiveness of the model trained by our simulation data and compared it with the models trained with manually labeled real data. These simulation and real data were randomly selected from the AADS-PC and ApolloScape-PC datasets, respectively. The mAP evaluation results of the instance segmentation models are presented in Fig. 6A. When trained with only our simulation data, the instance segmentation models produced results competitive with the precisely labeled real data. When using 100,000 data points generated by simulation, the segmentation performance was better than a model trained on 4000 real data points and came close to models trained on 16,000 and 100,000 real data points. In short, by using simulation to increase the size of the training set, performance can approach that of models trained on real-world data.

Next, we used simulation data to boost the real data (i.e., pretrain the model), as shown in Fig. 6C. Boosting with simulation data significantly improves (by 2 to 4%) the validation accuracy of the original model trained with only the real data. On the ApolloScape-PC dataset, we found that using 100,000 simulated data points to pretrain the model and 1600 real data for fine-tuning outperformed a model trained with 16,000 real data points in terms of the average mAP of all object types. When fine-tuned with 32,000 real data points, the model surpassed a model trained on 100,000 real data. These results indicate that our simulation approach may reduce up to 80 to 90% of manually labeled data, greatly reducing the need to label images and thus saving time and money. More details can be found in (31).

Last, on the basis of instance segmentation, we compared our object placement (traffic simulation) method with alternative placement strategies, e.g., placing object randomly or under specific rules (32). As shown in Fig. 6B, the accuracy of models trained with simulated data outperformed (by 4 to 7%) those trained with the other object placement strategies. The accuracy of models trained with our simulated data is close to that of a model trained on real data (gap of only 1 to 4%, depending on the application).

**TrafficPredict with AADS-TRAJ**

To evaluate the effectiveness of synthesized traffic, i.e., trajectories of cars, cyclist, and pedestrians, we adopted the TrafficPredict method Ma et al. (33) for quantitative evaluation. This method takes motion patterns of traffic agents in the first $T_{obs}$ frames as input and predicts their positions in the following $T_{pred}$ frames. In our evaluation, $T_{obs}$ and $T_{pred}$ were set to 5 and 7, respectively. We extended 20,000 real frames from ApolloScape-TRAJ dataset with an additional 20,000 simulated frames from our AADS-TRAJ dataset to train the deep neural network proposed in the method by Ma et al. (33). Performance of the trained model was measured using mean Euclidean distance between predicted positions and ground truth. In our case, average displacement error (mean Euclidean distance of all predicted frames) and final displacement error (mean Euclidean distance of the $T_{pred}$-th predicted frame) were evaluated. Prediction error was reduced sharply when training with an additional 20,000 simulated data points (Fig. 7). The error rate for cars was reduced the most because cars were well represented in the simulated trajectories.

## CONCLUSIONS

In the previous section, we showed the effectiveness of AADS for various tasks in AD. All of these tasks were achieved by using captured scene data (location specific) and traffic trajectory data (general). The entire AADS system requires very little human intervention. The system may be used to generate large amounts of realistic training data with fine annotation, or it may be used in-line to simulate the entire AD system from perception to planning. The realism and scalability of AADS make it possible to be used in real-world scenarios, as long as the background can be captured.

Compared to VR-based simulations, AADS's viewpoint change for RGB data is limited. Deviating too much from the original captured viewpoints leads to degraded image quality. However, we believe that the limited viewing range is actually acceptable for AD simulation. For the most part, a vehicle drives on flat roads, and the possible viewpoint changes are limited to rotation and 2D translation on the road plane. There is no need to support a bird's eye view or a third-person perspective for RGB-based perception. Another major limitation of AADS is the lack of lighting/environmental changes (snow/rain) in the scene. For now, these must be the same in the captured images,

| dataset | mAP(Bbox) | mAP(mask) |
|---|---|---|
| 16k sim | 91.02 | 72.63 |
| 4k real | 93.27 | 75.8 |
| 100k sim | 94.1 | 75.32 |
| 16k real | 94.41 | 79.21 |
| 100k real | 94.39 | 80.55 |

A

| methods | mAP(Bbox) | mAP(mask) |
|---|---|---|
| Random | 28.68 | 65.96 |
| Rule-based | 39.43 | 74.32 |
| Ours | 44.14 | 82.47 |
| Real data | 46.57 | 86.33 |

B

| dataset | car | pedestrian | cyclist | all |
|---|---|---|---|---|
| 100k sim | 91.02 | 72.63 | 57.21 | 83.03 |
| 16k real | 93.27 | 75.80 | 62.62 | 86.33 |
| 100k sim + 1.6k real | 94.10 | 75.32 | 61.35 | 86.40 |
| 100k sim + 16k real | 94.41 | 79.21 | 67.01 | 88.31 |
| 100k real | 94.39 | 80.55 | 67.30 | 88.50 |
| 100k sim + 32k real | 94.91 | 80.61 | 67.51 | 88.91 |
| 100k sim + 100k real | 95.27 | 81.20 | 68.25 | 89.30 |

C

**Fig. 6. LiDAR simulation evaluations.** (**A**) Evaluation of dataset's size and type (real or simulation) for real-time instance segmentation. (**B**) Evaluation results of different object placement methods. (**C**) Real data boosting evaluations (mean mask AP) using instance segmentation.

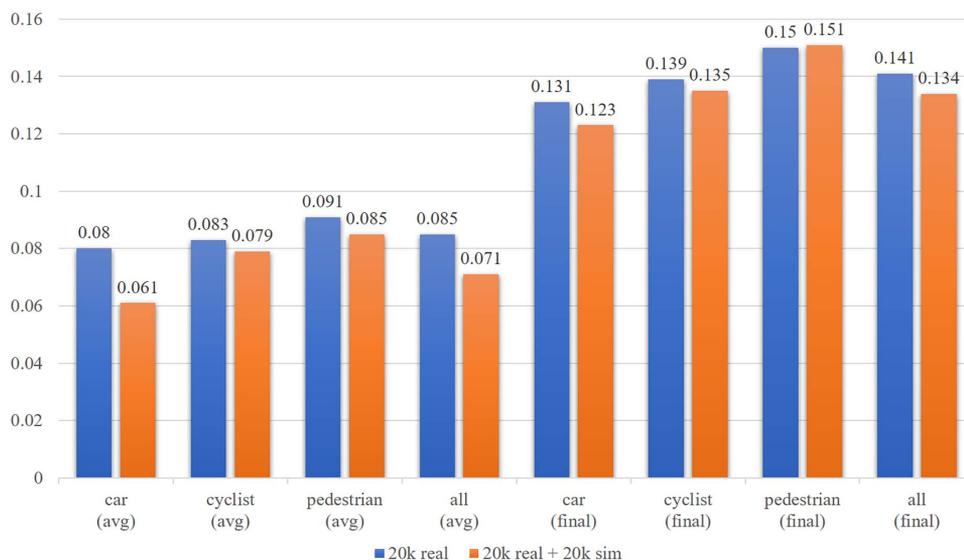

Fig. 7. **TrafficPredict evaluations.** Comparison of trajectory prediction with 20,000 real trajectory frames and an additional 20,000 simulation trajectory frames.

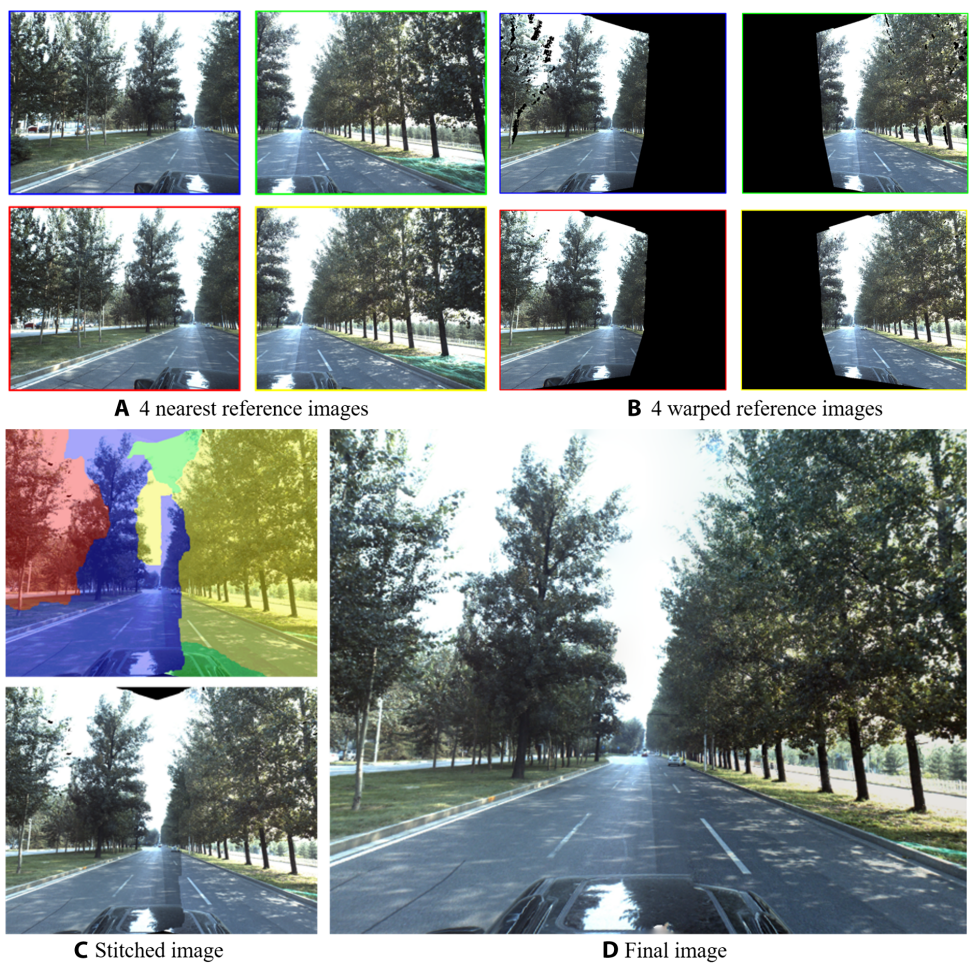

Fig. 8. **Novel view synthesis pipeline.** (**A**) The four nearest reference images were used to synthesize the target view in (D). (**B**) The four reference images were warped into the target view via depth proxy. (**C**) A stitching method was used to yield a complete image. (**D**) Final results in the novel view were synthesized after post-processing, e.g., hole filling and color blending.

but there have been significant advances in image synthesis using generative adversary networks (34, 35). Preliminary results synthesizing seasonal changes have been demonstrated. Enabling full lighting/environmental effect synthesis within AADS is a promising direction that we are actively pursuing.

## MATERIALS AND METHODS
### Data preprocessing
AADS used the scanned real images for simulation. Our goal was to simulate new vehicles and pedestrians in the scanned scene with new trajectories. To achieve this goal, before simulating data, our AADS should remove moving objects, e.g., vehicles and pedestrians, from scanned RGB images and point clouds. Automatic detection and removal of moving objects constitute a full research topic in its own right; fortunately, most recent datasets provide semantic labels of RGB images, including point clouds. By using semantic information in the ApolloScape dataset, we removed objects of a specific type, e.g., cars, bicycles, trucks, and pedestrians. After removing moving objects, numerous holes in both RGB images and point clouds appear, which must be carefully filled to generate a complete and clean background for AADS. We used the most recent RGB image inpainting method (36) to close the holes in the images. This method used the semantic label to guide a learning-based inpainting technique, which achieved acceptable levels of quality. The point cloud completion will be introduced in the depth processing for novel background synthesis (see the next section).

Given synthesized background images, we could place any 3D CG model on the ground and then render it into the image space to create a new, composite simulation image. However, to make the composite image photorealistic (look close to the real image), we must first estimate the illumination in background images. This enables our AADS to render 3D CG models with consistent shadows on the ground and on vehicle bodies. We solved such outdoor lighting estimation problems according to the method in (37). In addition, to further improve the reality of composite images, our AADS also provided an optional feature to enhance the appearance of 3D CG models by grabbing textures from real images. Specifically, given an RGB image with unremoved vehicles, we retrieved the corresponding 3D vehicle models and aligned those models to the input image using the method in (38). Similar to (39), we then used symmetric priors to transfer and complete the appearance of 3D CG models from aligned real images.

### Augmented background synthesis

Given a dense point cloud and image sequence produced from automatic scanning, the most straightforward way to build virtual assets for an AD simulator is to reconstruct the full environment. This line of work focuses on using geometry and texture reconstruction methods to produce complete large-scale 3D models from the captured real scene. However, these methods cannot avoid hand editing while modeling, which is expensive in terms of time, computation, and storage.

Here, we directly synthesized an augmented background in a specific view as needed during simulation. Our method avoided modeling the full scene ahead of running the simulation. Technically, our method created such a scan-and-simulate system by using the view synthesis technique.

To synthesize a target view, we needed to first obtain dense depth maps for input reference images. Ideally, these depth maps should be extracted from a scanned point cloud. Unfortunately, such depth maps will be incomplete and unreliable. In our case, these problems came with scanning: (i) The baseline of our stereo camera was too small compared to the size of street view scenes, and consequently, there were too few data points for objects that were too far from the camera. (ii) The scenes were full of numerous moving vehicles and pedestrians that needed to be removed. Unfortunately, their removal produced holes in scanned point clouds. (iii) The scenes were always complicated (e.g., many buildings are fully covered with glasses), which led to scanning sensor failure and thus incomplete scanned point clouds. We introduced a two-step procedure to address the lack of reliability and incompleteness in depth maps: depth filtering and depth completion.

With respect to depth filtering, we carried out a judiciously selected combination of pruning filters. The first pruning filter is a median filter: A pixel was pruned if its depth value was sufficiently different from the median filtered value. To prevent removing thin structures, the kernel size of the median filter was set to small (e.g., 5 by 5 in our implementation). Then, a guided filter (40) was applied to keep thin structures and to enhance edge alignment between the depth map and the color image. After getting a much more reliable depth, we completed the depth map by propagating the existing depth value to the pixels in the holes by solving a first-order Poisson equation similar to the one used in colorization algorithms (41).

After depth filtering and completion, reliable dense depth maps that could provide enough geometry information to render an image into virtual views were produced. Similar to (23), given a target virtual view, we selected the four nearest reference views to synthesize the virtual view. For each reference view, we first used the forward mapping method to produce a depth map with camera parameters of the virtual view and then performed depth inpainting to close small holes. Then, a backward mapping method and occlusion test were used to warp the reference color image into the target view.

A naïve way to synthesize the target image is to blend all the warped images together. However, when we blended the warped images using the view angle penalty following a previous work (23), there always existed obvious artifacts. Thus, we solved this problem as an image stitching problem rather than direct blending. Technically, for each pixel $x_i$ of the synthesized image in the target virtual view, it is optimized to choose a color from one of those warped images. This can be formulated as a discrete pixel labeling energy function

$$\arg\min_{\{x_i\}} \sum_i \lambda_1 E_1(x_i) + \lambda_2 E_2(x_i) + \sum_{(ij) \in N} \lambda_3 E_3(x_i, x_j) + \lambda_4 E_4(x_i, x_j) + \lambda_5 E_5(x_i, x_j) \quad (1)$$

Here, $x_i$ is the $i$th pixel of the target image, and $N$ is the pixel set of $x_i$'s one ring neighbor. $E_1(x_i)$ is the pixel-wise data term, which is defined by extending the view angle penalty in (42). In contrast to the scenarios in (42), depth maps of the street view scene always contain pixels with large depth values, which lead the angle view penalty to be too small. To address this problem, when the penalty is close to zero, we added another term to help choose the appropriate image by taking advantage of camera position information. Specifically, $E_1(x_i)$ is defined as $E_1(x_i) = E_{\text{angle}}(x_i) W_{\text{label}}(x_i)$. Here, $E_{\text{angle}}(x_i)$ is the view angle penalty in (42). When $E_{\text{angle}}(x_i)$ is too small, it will be hooked and set to 0.01 in our implementation. This is done to balance two energy terms and make $W_{\text{label}}(x_i)$ effective. $W_{\text{label}}(x_i)$ is defined as $W_{\text{label}}(x_i) = D_{\text{pos}}(C_{x_i}, C_{\text{syn}}) D_{\text{dir}}(C_{x_i}, C_{\text{syn}})$, which evaluates the difference between the reference view and the target view. Here, $C_{x_i}$ and $C_{\text{syn}}$ denote the view choice for the camera for pixel $x_i$ and for the target view's camera, respectively. Furthermore, $D_{\text{pos}}$ represents the distance from the camera center, and $D_{\text{dir}}$ is the angle between the optical axes of the two cameras.

$E_2(x_i)$ is the occlusion term used to exclude the occlusion areas while minimizing the pixel labeling energy. Most occlusions appear near depth edges. Thus, when using the backward mapping method to render the warped images, we detected occlusions by performing depth testing. All pixels in the reference view with larger depth values than those of the source depth can yield an occlusion mask, which is then used to define $E_2(x_i)$. Specifically, when an occlusion mask is invalid, i.e., the pixel is nonocclusion, we set $E_2(x_i) = 0$ to add no penalty into the energy function. When a pixel is occluded, we set $E_2(x_i) = \infty$ to exclude this pixel completely.

The rest of the terms in Eq. 1 are smoothness terms: color term $E_3(x_i, x_j)$, depth term $E_4(x_i, x_j)$, and color gradient term $E_5(x_i, x_j)$. Similar to (43), the color term $E_3(x_i, x_j)$ is defined by a truncated seam-hiding pairwise cost first introduced in (44) $E_3(x_i, x_j) = \min(||c_i^{x_i} - c_i^{x_j}||^2, \tau_c) + \min(||c_j^{x_i} - c_j^{x_j}||^2, \tau_c)$, where $c_i^{x_i}$ is the RGB value of pixel $x_i$. Similarly, the depth term $E_4(x_i, x_j)$ is defined as $E_4(x_i, x_j) = \min(|d_i^{x_i} - d_i^{x_j}|, \tau_d) + \min(|d_j^{x_i} - d_j^{x_j}|, \tau_d)$, where $d_i^{x_i}$ is the depth of pixel $x_i$, and the RGB and depth truncation thresholds are set to $\tau_c = 0.5$ and $\tau_d = 5$ m in our implementation, respectively. Because the illumination difference may occur between different reference images, the color difference is not sufficient to ensure a good stitch. An additional gradient difference $E_5(x_i, x_j)$ is used. By assuming that the gradient vector should be similar on both sides of the seam, we defined $E_5(x_i, x_j) = |g_i^x - g_j^x| + |g_i^y - g_j^y|$, where $g_i^x$ is a color space gradient of the $i$th pixel in the image and includes $x_i$.

The term weights in Eq. 1 are set to $\lambda_1 = 200$, $\lambda_2 = 1$, $\lambda_3 = 200$, $\lambda_4 = 100$, and $\lambda_5 = 50$. The labeling problems are solved using the sequential tree-reweighted message passing (TRW-S) method (45). Figure 8 shows the pipeline and results of augmented background synthesis. Note that, in Fig. 8C, a color difference may exist near the stitching seams after image stitching. To obtain consistent results, a modified Poisson image blending method (46) was performed. Specifically, we selected the nearest reference image as the source domain and then fixed its edges to propagate color brightness to the other side of the stitch seams. After solving the Poisson equation, we obtained the fusion result shown in Fig. 8D. Note that, when the novel view is far from the input views, e.g., large view extrapolation, there will be artifacts because of disoccluded regions that cannot be filled in by stitching together pieces of the input images. We marked those regions as holes and set their gradient value to zero. Thus, these holes were filled with plausible blurred color when solving the Poisson equation.

### Moving objects' synthesis and data augmentation

With a synthesized background image in the target view, a complete simulator should have the ability to synthesize realistic traffics with diverse moving objects (e.g., vehicles, bicycles, and pedestrians) and produce corresponding semantic labels and bounding boxes in simulated images and LiDAR point clouds.

We used the data-driven method described in (26) to address challenges involving traffic generation and placement of moving objects. Specifically, given the localization information, we first extracted lane information from an associated high-definition (HD) map. Then, we randomly initialized the moving objects' positions within lanes and ensured that the directions of moving objects were consistent with the lanes. We used agents to simulate objects' movements under constraints such as avoiding collisions and yielding to pedestrians. The multiagent system was iteratively deduced and optimized using previously captured traffic following a data-driven method. Specifically, we estimated motion states from our real-world trajectory dataset ApolloScape-TRAJ; these motion states included position, velocity, and control direction information of cars, cyclists, and pedestrians. Note that such real dataset processing was performed in advance of simulation and needed to be processed just once. During simulation runtime, we used an interactive optimization algorithm to make decisions for each agent at each frame of the simulation. In particular, we solved this optimization problem by choosing a velocity from the datasets that tends to minimize our energy function. The energy function was defined on the basis of the locomotion or dynamics rules of heterogeneous agents, including continuity of velocity, collision avoidance, attraction, direction control, and other user-defined constraints.

With generated traffic, i.e., the object placement in each simulation frame, we rendered 3D models into the RGB image space and generated annotated data using the physical rendering engine PBRT (47). Meanwhile, we also generated a corresponding LiDAR point cloud with annotations using the method introduced in the next section.

### LiDAR synthesis

Given 3D models and corresponding placement, it is relatively straightforward to synthesize LiDAR point clouds with popular simulators such as CARLA (4). Nevertheless, there are opportunities to take advantage of specific LiDAR sensors (e.g., Velodyne HDL-64E S3). We proposed a realistic point cloud synthesis method by effectively modeling the specific LiDAR sensor following a data-driven fashion. Technically, a real LiDAR sensor captured the surrounding scene by measuring the time of flight for pulses of each laser beam (48). One laser beam was emitted from the LiDAR and then reflected from target surfaces. A 3D point was then generated if the returned pulse energy of a laser beam was big enough. We modeled the behavior of laser beams to simulate this physical process. Specifically, the emitted laser beam could be modeled using parameters including the vertical and azimuth angles and their angular noises, as well as the distance measurement noise. For example, the Velodyne HDL-64E S3 LiDAR sensor emits 64 laser beams in different vertical angles ranging from −24.33° to 2°. During data acquisition, HDL-64E S3 rotates around its own upright direction and shoots laser beams at a predefined rate to accomplish 360° coverage of the scenes. Ideally, such models can be adaptive to other types of LiDAR sensors. Model parameters should depend on the specific type of sensor. However, we experimentally found that parameters vary considerably, even among devices of the same type. To be as close as possible to reality, we fitted the model from real point clouds to statistically derive those parameters.

Specifically, we collected real point clouds from HDL-64E S3 sensors on top of parked vehicles, guaranteeing smoothness of point curves from different laser beams. The points of each laser beam were then marked manually and fitted by a cone with the apex located in the LiDAR center. The half-angle of the cone minus $\pi/2$ forms the real vertical angle, whereas the noise variance was determined from the deviation of lines constructed by the cone apex and points from the cone surface. The real vertical angles usually differed from ideal angles by 1° to 3°. In our implementation, we modeled the noise with standard Gaussian distribution, setting the distance noise variance to 0.5 cm and the azimuth angular noise variance to 0.05°.

To generate a point cloud, we computed intersections between the laser beams and the scene. Specifically, we proposed a cubed, map-based method to mix the background of the scenes in the form of points and meshes of 3D CG models. Instead of computing intersections between the beams and the mixed data, we computed the intersection with the projected maps (e.g., depth map) of scenes, which offer the equivalent information but in a much simpler form. Note that our LiDAR simulation method can be easily extended for arbitrary LiDAR sensors and to any sensor solution for different numbers and poses of sensors. Figure S1 shows the visual results of our LiDAR simulation.

### SUPPLEMENTARY MATERIALS
robotics.sciencemag.org/cgi/content/full/4/28/eaaw0863/DC1
Fig. S1. Visual evaluations of point cloud simulation.
Movie S1. Full movie.
Movie S2. Scan-and-simulation pipeline.
Movie S3. Synthesizing lane changes.
Movie S4. Data augmentation.
Movie S5. Novel view synthesis evaluations.

**Funding:** This work was supported by NSFC grants 61732016 (to W.W.X.), 61872398 and 61632003 (to G.P.W.), and National Key R&D Program of China 2017YFB1002700 (to G.P.W.). **Author contributions:** R.G.Y. conceived the project. W.L. and C.W.P. developed the concept and systems. J.P.R. developed the trajectory synthesis framework. R.Z. and Q.C.G. performed the synthesized RGB image evaluations. X.Y.H. helped collect the RGB and point cloud datasets. J.F. and F.L.Y. performed the synthesized LiDAR point cloud evaluations. Y.X.M. helped collect the trajectories dataset and performed the simulated trajectory evaluations. G.P.W., W.W.X., and H.J.G. discussed the results and contributed to the final manuscript. W.L., D.M., and R.G.Y. wrote the paper. **Competing interests:** W.L., C.W.P., and R.Z. completed the work while interning with Baidu Research. F.L.Y., J.F., and R.G.Y. are inventors on patent application no. CN20181105574.2, "A method of LiDAR point cloud simulation for autonomous driving." The other authors declare that they have no competing interests. **Data and materials availability:** The RGB and point cloud datasets (ApolloScape-RGB and ApolloScape-PC) are hosted with the web link http://apolloscape.auto/scene.html. The trajectory dataset (ApolloScape-TRAJ) announced along with this paper can be freely downloaded through the link http://apolloscape.auto/trajectory.html. Some of the data used and the code are proprietary.

Submitted 4 December 2018
Accepted 5 March 2019
Published 27 March 2019
10.1126/scirobotics.aaw0863

**Citation:** W. Li, C. W. Pan, R. Zhang, J. P. Ren, Y. X. Ma, J. Fang, F. L. Yan, Q. C. Geng, X. Y. Huang, H. J. Gong, W. W. Xu, G. P. Wang, D. Manocha, R. G. Yang, AADS: Augmented autonomous driving simulation using data-driven algorithms. *Sci. Robot.* **4**, eaaw0863 (2019).

# Science Robotics

**AADS: Augmented autonomous driving simulation using data-driven algorithms**

W. Li, C. W. Pan, R. Zhang, J. P. Ren, Y. X. Ma, J. Fang, F. L. Yan, Q. C. Geng, X. Y. Huang, H. J. Gong, W. W. Xu, G. P. Wang, D. Manocha and R. G. Yang